\documentclass{article}
\usepackage{enumitem}
\setlist{nosep, leftmargin=14pt}

\usepackage{graphicx}
\usepackage[dvipsnames]{xcolor}
\usepackage{multirow}
\usepackage{subfigure}
\usepackage{makecell}
\usepackage{soul}
\usepackage{amsmath}
\usepackage{siunitx}
\usepackage{amsfonts}
\usepackage[dvipsnames]{xcolor}
\usepackage{appendix}
\usepackage[preprint]{spconf}
\usepackage{float}
\usepackage{hyperref}
\toappear{\textit{Preprint accepted at Computerized Medical Imaging and Graphics}}

\title{Improving Vessel Segmentation with Multi-Task Learning and Auxiliary Data Available Only During Model Training}

\name{\begin{tabular}{c}Daniel Sobotka$^{1,*\thanks{* Equal contribution}}$, Alexander Herold$^{2,*}$, Matthias Perkonigg$^{1,3}$, Lucian Beer$^{2}$, Nina Bastati$^{2}$, \\ Alina Sablatnig$^{1}$, Ahmed Ba-Ssalamah$^{2}$ and Georg Langs$^{1}$\end{tabular}}

\address{$^{1}$ Computational Imaging Research Lab , Department of Biomedical Imaging and  Image-guided Therapy, \\ Medical University of Vienna, Vienna, Austria \\
$^{2}$ Division of General and Paediatric Radiology, Department of Biomedical Imaging \\ and Image-guided Therapy, Medical University of Vienna, Vienna, Austria \\
$^{3}$ Department of Medical Statistics, Informatics and Health Economics, Medical University of Innsbruck, \\ Innsbruck, Austria}

\begin{document}
\maketitle
\begin{abstract}
Liver vessel segmentation in magnetic resonance imaging data is important for the computational analysis of vascular remodelling, associated with a wide spectrum of diffuse liver diseases. Existing approaches rely on contrast enhanced imaging data, but the necessary dedicated imaging sequences are not uniformly acquired. Images without contrast enhancement are acquired more frequently, but vessel segmentation is challenging, and requires large-scale annotated data. We propose a multi-task learning framework to segment vessels in liver MRI without contrast. It exploits auxiliary contrast enhanced MRI data available only during training to reduce the need for annotated training examples. Our approach draws on paired native and contrast enhanced data with and without vessel annotations for model training. Results show that auxiliary data improves the accuracy of vessel segmentation, even if they are not available during inference. The advantage is most pronounced if only few annotations are available for training, since the feature representation benefits from the shared task structure. A validation of this approach to augment a model for brain tumor segmentation confirms its benefits across different domains. An auxiliary informative imaging modality can augment expert annotations even if it is only available during training. 
\end{abstract}
\section{Introduction}

\label{sec:introduction}
Contrast enhanced magnetic resonance imaging (MRI) is used to assess fine grained soft tissue properties when studying diffuse liver diseases and their progress to liver cirrhosis. The difficulty of standardization, the risk and time constraints associated with contrast enhanced imaging \cite{khawaja2015revisiting} limit the range of sequences typically acquired in protocols optimized for specific diagnostic tasks and efficiency. 

\subsection{Clinical context} 

Automatic liver vessel segmentation using machine learning offers a means to study vascular remodelling associated with a wide spectrum of diffuse liver diseases, such as (non-)alcoholic steatohepatitis, \cite{scaglioni2011ash} or (non-)alcoholic fatty liver disease \cite{lakshman2015synergy}, in large cohorts. It is also crucial for planning liver resection of primary liver tumour, such as hepatocellular carcinoma (HCC) complicating chronic liver diseases (CLD), or rarely metastases in the cirrhotic liver \cite{alirr2023hepatic}. Vessel segmentation can also be used to assess the vasculature of the living donor before liver transplantation for HCC or liver failure \cite{alirr2018automated}.
\textcolor{Black}{For} instance, the link between vessel alterations and chronic liver disease progression or their relationship to other noninvasive imaging markers such as functional liver imaging score \cite{bastati2020does} are subject of research. However, large scale studies that would benefit from computational quantification have to rely on frequently acquired sequences without contrast enhancement, rendering vessel segmentation difficult. This reflects a pattern of availability of more sophisticated imaging modalities for smaller data sets, and the need to apply machine learning models to larger case numbers for which fewer imaging modalities are available. The present paper proposes an approach to exploit such \emph{auxiliary} modalities during training of models, even if they are not available at inference time. 

\begin{figure*}[t]
\centerline{\includegraphics[width=\textwidth]{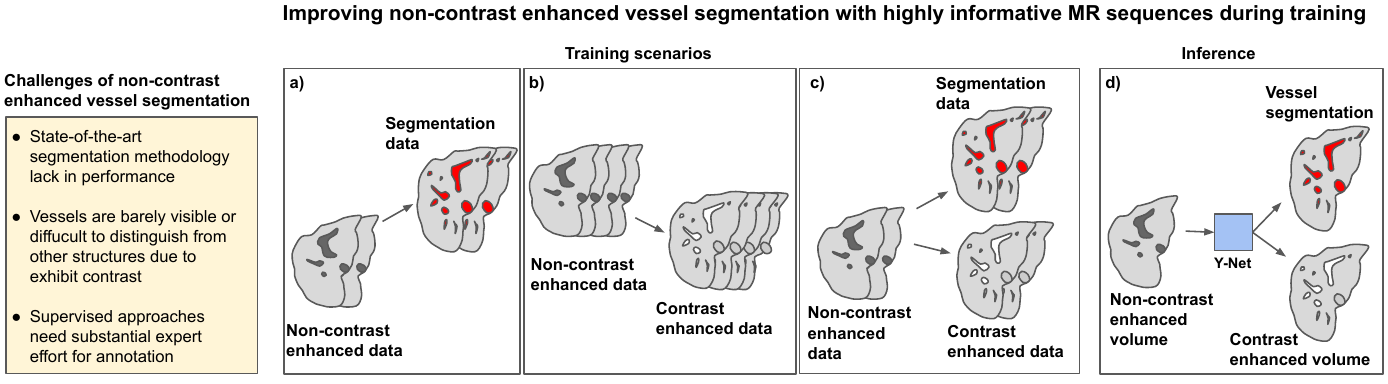}}
\caption{Overview of the three training data scenarios (a,b,c) and inference (d): (a) Pairs of non-contrast enhanced imaging data with corresponding vessel annotations; (b) pairs of non-contrast enhanced and contrast enhanced images, but without annotations; (c) triplets of corresponding non-contrast enhanced images, vessel segmentations and contrast enhanced images; (d) after training the model predicts vessel segmentation and contrast enhanced image from a non-contrast enhanced image.} 
\label{fig:data_overview}
\end{figure*}

\subsection{State of the art in liver vessel segmentation} 

Methods for liver vessel segmentation in imaging data can be divided into four groups \cite{ciecholewski2021computational}: (1) vessel enhancement filtering~\cite{frangi1998multiscale,sato1998three,meijering2004design,soares2006retinal}, (2) active contours~\cite{lu2017hepatic,zeng2018automatic,chung2018accurate}, (3) vessel tracking~\cite{lebre2019automatic,guo2020novel} and (4) machine learning methods~\cite{ibragimov2017combining,huang2018robust,mishra2018segmentation, kitrungrotsakul2019vesselnet,keshwani2020topnet,thomson2020mr,yan2020attention,xu2020training}. Approaches (1)-(3) are sensitive to image intensity changes or edges and work well for segmenting vessels in contrast enhanced data if there is substantial effort in post-processing \cite{marcan2014segmentation}. These methods fair worse in MRI data without contrast enhancement, where vessels are often barely visible, or imaging artifacts are present~\cite{goceri2017vessel}. To some extent, machine learning models, either unsupervised or supervised, can cope with this problem. However, supervised methods require a substantial amount of annotated training data to learn a mapping from complex imaging data to vessel label maps. Recent research \cite{huang2018robust,thomson2020mr} exploits a 3D U-Net \cite{cciccek20163d} for liver vessel segmentation in contrast enhanced MRI and CT images. 

\subsection{State of the art in multi-task learning (MTL)} 

MTL aims to improve generalization performance of multiple related tasks by leveraging useful shared information of these tasks \cite{caruana1997multitask,zhang2021survey}. In contrast to transfer learning, where the performance of a target task is improved with the help of a source task, MTL treats all tasks equally \cite{zhang2021survey}. Assuming task compatibility, the inclusion of an additional task to the MTL network can improve the performance of the model for the initial task, even if the accuracy on the new task is poor \cite{standley2019tasks}. For instance, image segmentation and image reconstruction harmonize in medical image MTL applications \cite{weninger2019multi,amyar2020multi}. Different data sets for each task can improve an effective shared feature representation~\cite{amyar2020multi,zhang2017survey}. MTL models can be grouped into encoder-focused models (EFM) and decoder-focused models (DFM), that either share hard- (shared and task specific parameters) or soft-parameters (tasks specific parameters with cross-talk such as cross-stitch units \cite{misra2016cross})~\cite{vandenhende2020revisiting}. In EFM and DFM all tasks share an encoder but, EFM uses its own decoder for each task and DFMs first predict initial task predictions and then leverage them to improve the task output \cite{vandenhende2020revisiting}. Recent approaches have addressed task balancing using task uncertainty \cite{kendall2018multi}, gradient normalization \cite{chen2018gradnorm}, dynamic task prioritization \cite{guo2018dynamic} or dynamic weight averaging \cite{liu2019end} to further enhance these models. Multi-task learning approaches with a shared decoder and different encoders for semantic segmentation, image reconstruction and classification tasks used for lung lesion segmentation \cite{amyar2020multi}, brain tumor segmentation \cite{weninger2019multi}, or left atrial segmentation and scar quantification in the heart \cite{li2020joint}. 

\subsection{Contribution} \textcolor{Black}{This paper introduces a semi-supervised multi-task machine learning approach to enhance image segmentation using privileged, \emph{auxiliary} multi-modal medical imaging data during the training phase (Fig.\,\ref{fig:data_overview}). A Y-net model learns to simultaneously predict the segmentation target (vessel labels), and the closely related auxiliary imaging modality (contrast enhanced imaging) from a source modality (non-contrast imaging) using one encoder, and two decoders linked with neural discriminative dimensionality reduction layers. This improves segmentation accuracy from the source modality, even if the auxiliary modality is not available during test time. Experiments show that this approach is particularly effective if only few annotated training examples are available, since the model even benefits from unannotated paired source- and  auxiliary modality examples. Evaluation results demonstrate the benefits of the technique on two data sets, including liver vessel segmentation, and brain tumor segmentation from non-contrast MRI.}

\section{Methodology}

We propose a shared encoder multi-task learning approach for a 3D U-Net architecture \cite{cciccek20163d} for semantic segmentation. The network consists of a Y-Net~\cite{wang2019net} with a shared encoder and task specific decoders, where one input image volume $\bf{I}$ is mapped to an output vessel label volume $\bf{L}$ and the auxiliary contrast-enhanced image volume $\bf{J}$, $\bf{I}\mapsto \langle \bf{L},\bf{J} \rangle$.  
Layer-wise feature fusion \cite{gao2019nddr} is used to share features between the two task specific decoders. We assume to have three types of data, non-contrast enhanced MRI T1w data $\bf{I}_i$, and contrast enhanced MRI T1wce data $\bf{J}_i$ with $i=1,\dots,n$. For a sub-set $i=1,\dots,m<n$ we also have expert vessel label annotations in the form of binary image volumes $\bf{L}_i$. During segmentation of new data, a non-contrast enhanced T1w image $\mathbf{I}_{new}$ serves as input and the model predicts vessel labels $\hat{\mathbf{L}}_{new}$, and an estimate of a contrast enhanced image $\hat{\mathbf{J}}_{new}$. $\bf{J}_i$ and $\bf{L}_i$  are used only during training. 
\begin{figure}[t]
\centerline{\includegraphics[width=\columnwidth]{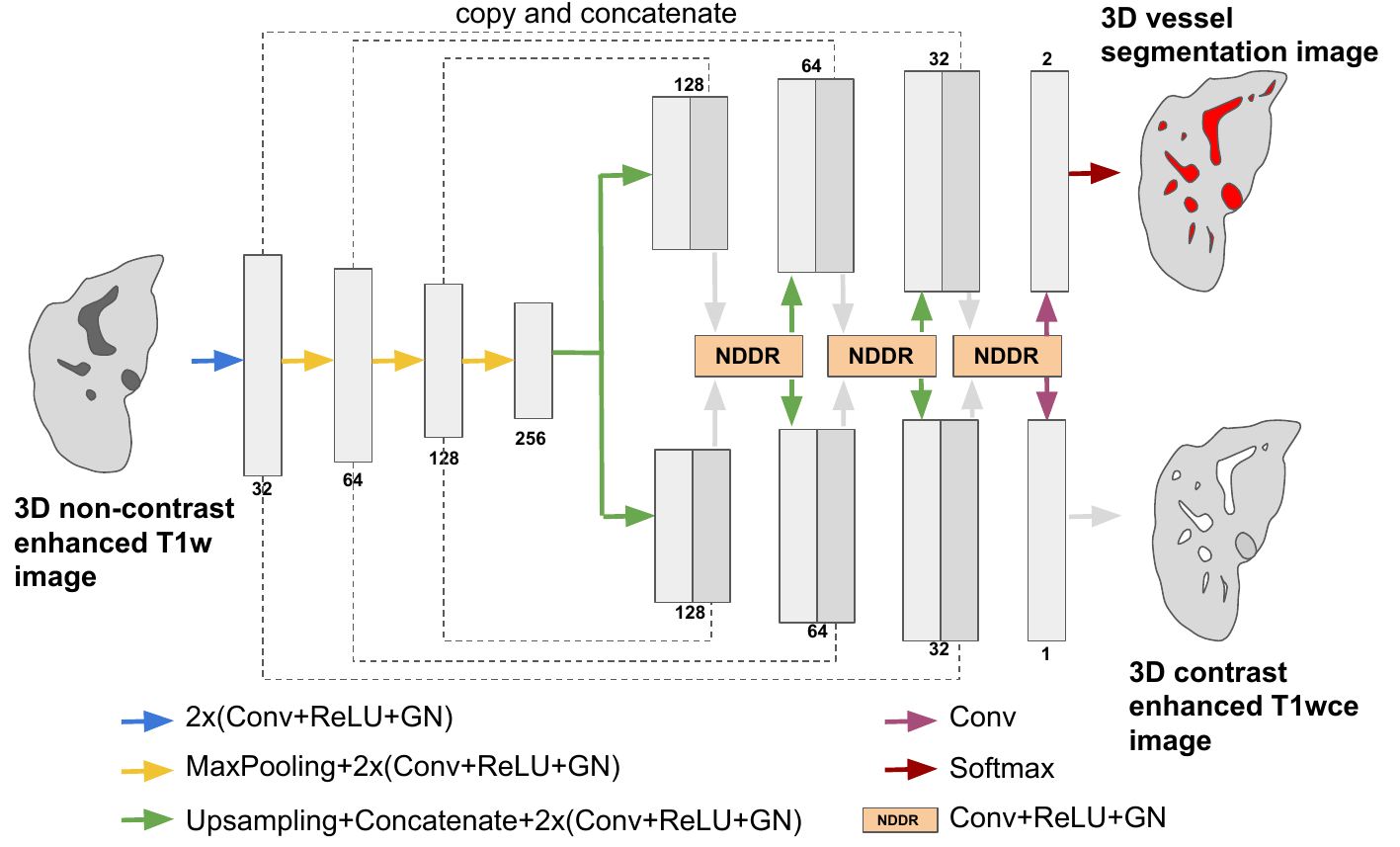}}
\caption{The architecture of the proposed multi-task framework with a shared encoder and a decoder for each task $T_{S}$, $T_{T}$. The input of the network is a 3D non-contrast enhanced T1w image and the outputs are the 3D vessel segmentation ($T_{S}$) and the translated 3D contrast enhanced T1wce image ($T_{T}$).} 
\label{fig:arch}
\end{figure}

\subsection{A Y-Net segments vessels and predicts contrast enhancement}
We train a Y-net encoder consisting of three downsampling blocks, and two decoders of three upsampling blocks each (Fig. \ref{fig:arch}). Each downsampling block contains two $3\times3\times3$ convolutions followed by Rectified Linear Units (ReLU) and Group Normalizations (GN) \cite{wu2018group}, as well as a $2\times2\times2$ max pooling operation. The upsampling path for both, segmentation and image translation tasks ($T_{S}$ and $T_{T}$, resp.) is symmetric to the downsampling path and uses nearest neighbor interpolation. To utilize features learned from the two tasks, we use Neural Discriminative Dimensionality Reduction (NDDR) layers \cite{gao2019nddr} to learn the optimal structure for layerwise feature fusing. Therefore, features with the same spatial resolution from both decoders are concatenated followed by a $1\times1\times1$ convolution, ReLu and GN. The loss function $L$ balances the single-task losses for the weights $W$ of the network using homoscedastic uncertainty \cite{kendall2018multi}:
\begin{equation}
\begin{split}
L(W,\sigma_{S},\sigma_{T}) = \frac{1}{2\sigma_{S}^2}L_{S}(W) + \frac{1}{2\sigma_{T}^2}L_{T}(W) + \log{(\sigma_{S})} \\
+ \log{(\sigma_{T})},
\end{split}
\end{equation}
where $L_{S}(W)$ denotes the cross entropy loss for $T_{S}$. $L_{T}(W)$ is the mean squared error loss for $T_{T}$ and $\sigma_{S}$ and $\sigma_{T}$ are parameters that are learned,  balancing the task-specific losses $L_{S}(W)$ and $L_{T}(W)$ during training. Lower $\sigma$ values increase the contribution of the specific task, where higher $\sigma$ values decrease its contribution. 

\subsection{Model training and inference}
\begin{figure*}[t]
\centerline{\includegraphics[width=\textwidth]{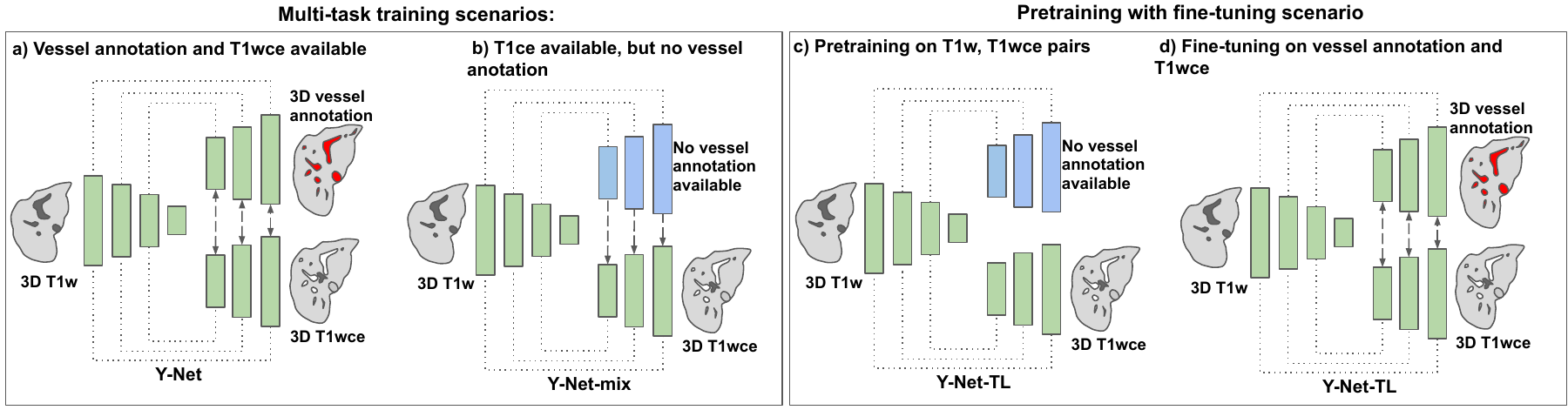}}
\caption{Training the multi-task learning model: a) multi-task training with paired data consisting of triplets of T1w, T1wce and corresponding vessel segmentations for each example, b) multi-task training on additional paired data of T1w / T1wce pairs without vessel annotation. Pretraining with fine-tuning: c) pretraining with T1w / T1wce pairs without vessel annotation, d) fine-tuning with triplets of T1w, T1wce and corresponding vessel segmentations. In section \ref{sec:exp1} and \ref{sec:exp3} the Y-Net is trained with a), while in section \ref{sec:exp2} the Y-Net mix is trained on a) and b). In section \ref{sec:exp3} the Y-Net-TL is trained with c) and d).}
\label{fig:train}
\end{figure*}
We explored two multi-task learning training scenarios. \textbf{(1)} Training is performed with triplets of $\langle \mathbf{I},\mathbf{J},\mathbf{L}\rangle$. \textbf{(2)} Training is performed with triplets of $\langle \mathbf{I},\mathbf{J},\mathbf{L}\rangle$ and additional pairs of $\langle \mathbf{I},\mathbf{J}\rangle$ without annotation. 
If the multi-task network is trained with triplets (Fig.~\ref{fig:train}a) both connected decoders with the shared encoder are trained simultaneous. If additional pairs of non-contrast enhanced and contrast enhanced images are available (Fig. \ref{fig:train}b) the layerwise feature fusion is used only in the direction of the image translation task and only the image translation decoder and the shared encoder are trained. In that case only the task specific loss of $L_{T}$ is evaluated. Additionally, we evaluate if initial transfer learning can further improve the model. To this end, the Y-Net is pretrained with T1w/T1wce pairs (Fig. \ref{fig:train}c), where only the shared encoder and the task specific $T_{T}$ decoder is trained. Second, fine-tuning (Fig. \ref{fig:train}d) is done with $\langle \mathbf{I},\mathbf{J},\mathbf{L}\rangle$ triplets. 

During inference a non-contrast enhanced T1w image $\mathbf{I}_{new}$ serves as input and the model predicts $\hat{\mathbf{L}}_{new}$ and $\hat{\mathbf{J}}_{new}$.

\section{Experiments}\label{sec:Results}

We evaluated if multi-task learning can improve the accuracy of segmentation models by comparing the performance of different training strategies to exploit privileged data available only during training to models trained only on data also available during test time.

\setlength{\tabcolsep}{3pt}

\begin{table*}[t]
  \begin{center}
    \caption{\textcolor{Black}{Learnable parameters, activation function and optimization function of all models. Note that additional parameters for Y-nets correspond to the auxiliary decoder arm, only used during training.}}
    \label{tab:model_details}
    \begin{tabular}{l | c | c | c}
      Model & Learnable parameters & Activation function & Optimization function \\ \hline
      U-Net & 4080914 & ReLu & Adam\\ 
      Y-Net & 6491189 & ReLu & Adam\\ 
      Y-Net-Mix & 6491187 & ReLu & Adam\\ 
      Y-Net-TL & 6491189 & ReLu & Adam\\ 
    \end{tabular}
  \end{center}
\end{table*}

\setlength{\tabcolsep}{3pt}
\begin{table*}[t]
  \begin{center}
    \caption{Segmentation accuracy (and standard deviation between the datasets) with $12$ triplets used for training between the proposed Y-Net and the baseline U-Net}
    \label{tab:exp1_seg}
    \begin{tabular}{l | c c c c c c }
      Model & Dice Score & Jaccard Score & Mean Surface Distance & Recall & Precision & \textcolor{Black}{Absolute VVR Difference}\\ \hline
      \textcolor{Black}{Frangi with Otsu} & 0.020 (0.01) & 0.010 (0.01) & 75.822 (10.455) & 0.225 (0.08) & 0.010 (0.01) & 2.293  \\
      U-Net & 0.446 (0.12) & 0.294 (0.10) & 4.615 (2.22) & 0.347 (0.13) & 0.703 (0.13) & 2.378 \\ 
      Y-Net & 0.506 (0.09) & 0.343 (0.08) & 3.929 (1.41) & 0.418 (0.10) & 0.679 (0.15) & 2.012 \\
    \end{tabular}
  \end{center}
\end{table*}

\setlength{\tabcolsep}{3pt}
\begin{table*}[h]
  \begin{center}
    \caption{Segmentation accuracy for Y-Net-mix with additional T1w/T1wce training pairs without annotation. All segmentation results are for applying the model to non-contrast enhanced T1w data.} 
    \label{tab:exp2_seg}
    \begin{tabular}{l | c c c c c c}
      Model & Dice Score & Jaccard Score & Mean Surface Distance & Recall & Precision & \textcolor{Black}{\makecell{Absolute\\VVR Difference}}\\
      \hline
      Y-Net-mix (5 T1w/T1wce) & 0.527 (0.09)  & 0.363 (0.09) & 3.409 (1.13) & 0.470 (0.11) & 0.631 (0.14) & 1.635 \\
      Y-Net-mix (10 T1w/T1wce) & 0.532 (0.11) & 0.370 (0.10) & 4.109 (1.71) & 0.464 (0.13) & 0.667 (0.15)& 1.848 \\
      Y-Net-mix (20 T1w/T1wce) & 0.545 (0.12) & 0.383 (0.11) & 3.985 (2.32) & 0.477 (0.14) & 0.681 (0.11) & 1.681 \\   
      Y-Net-mix (30 T1w/T1wce) & 0.539 (0.10) & 0.376 (0.10) & 3.621 (1.64) & 0.476 (0.12) & 0.670 (0.16) & 1.778 \\   
    \end{tabular}
  \end{center}
\end{table*}

\subsection{Data and Model}

We studied MR image volumes of $45$ chronic liver disease patients. For each patient a T1w image volume with (T1wce) and without contrast enhancement (T1w) was available. MRI was performed using a 3T MRI scanner (MRI Magnetom Trio, A Tim, Siemens Healthcare) with a combined six-element, phased-array abdominal coil and a fixed spine coil. Both, the T1w and the T1wce (portalvenous phase at $70$ seconds) were T1-weighted, volume-interpolated, breath-hold examination sequences with in-plane voxel spacing between $0.781$ and $0.8789$ millimeter and out-of-plane voxel spacing between $1.7$ and $2.3$ millimeter. Written informed consent was obtained from all patients and the study protocol was approved by the local ethics committee. For $15$ patients, hepatic veins and portal vein were segmented by radiologists on the T1wce image, and segmentation together with T1wce were registered to the T1w image. This resulted in $15$ $\langle \mathbf{I},\mathbf{J},\mathbf{L}\rangle$ triplets, and $30$ $\langle \mathbf{I},\mathbf{J}\rangle$ pairs. All the image volumes were Z-score normalized, resampled to $256\times192\times128$ for training and prediction and no further pre-processing or post-processing was performed.

\textcolor{Black}{Details on the models are provided in Table\,\ref{tab:model_details}. We used ReLu activation functions, and Adams optimizer. The Y-nets have $6491187$  trainable parameters compared to $4080914$ for U-nets. However, the additional parameters stem from the decoder arm to the auxiliary image modalities, and are thus only used during training, but not during the application of the model to new data. Each downsampling block contains two 3×3×3 convolutions followed by Rectified Linear Units (ReLU) and Group Normalizations (GN). All networks were optimized using Adam optimizer with an initial learning rate of 0.001
}

\subsection{Experiment set-up}\label{sec:exp}
We conducted five experiments. 
\textbf{(1)} In the first experiment we evaluated if an auxiliary image translation task during training can improve vessel segmentation accuracy. We evaluated segmentation accuracy using $15$ $\langle \mathbf{I},\mathbf{J},\mathbf{L}\rangle$ triplets of non-contrast enhanced T1w, contrast enhanced T1wce MRI data, and vessel annotations (Fig. \ref{fig:train}a) reporting Dice coefficient, Jaccard similarity coefficient, mean surface distance, recall and precision of the segmentation. \textcolor{Black}{Further, we introduced the Vessel-to-Volume ratio (VVR) as a metric:
\begin{equation}
    VVR = \frac{L_{v}}{V_{v}} 
\end{equation}
where $L_{v}$ denotes the liver volume and $V_{v}$ the corresponding vessel volume
and reported the absolute VVR difference between ground truth and predictions.}
\textcolor{Black}{We} performed $3$-fold cross-validation on a set of $12$ patients, in each fold training with $8$ triplets, and testing with $4$ triplets. We compared the 3D Y-Net with a 3D U-Net trained analogously to one arm of the 3D Y-Net, with only a single decoder without NDDR layers~(Fig.~\ref{fig:arch}) 
\textbf{(2)} We evaluated if additional paired examples without annotation $\langle \mathbf{I},\mathbf{J}\rangle$ can improve vessel segmentation accuracy further (Fig.~\ref{fig:train}b). For this $5$ to $30$ $\langle \mathbf{I},\mathbf{J}\rangle$ pairs without annotation were added to the training set, and segmentation accuracy was evaluated.
\textbf{(3)} We evaluated the segmentation accuracy using transfer learning between images without and with annotations for training. First, $5$ to $30$ $\langle \mathbf{I},\mathbf{J}\rangle$ pairs without annotation are used as pretraining and the T1w, T1wce and vessel annotation triplets are used only for fine-tuning. 
\textbf{(4)} In the fourth experiment the segmentation accuracy in relation to the number of annotations used during training is assessed. 
\textbf{(5)} Finally, we evaluated the segmentation accuracy of each experiment best model with the hold-out test dataset of $3$ cases which were not used during training or cross-validation for the entire segmented vessel tree and compared accuracy for vessels with different thickness.

Finally, we evaluated these models with different amounts of available training data by varying the number of annotated cases used, and for different vessel diameters. For the latter, we divided the ground truth vessel segmentations into $4$ vessel thickness groups from $0-5$mm, $5-10$mm, $10-15$mm and $>15$mm based on 3D distance maps. Predicted labels were assigned to these $4$ groups for Dice score evaluation. False positive labels were assigned to the nearest 3D ground truth skeleton voxel \cite{lee1994building}, and the vessel diameter from that voxel's label was used for referring the false positive label to the corresponding group. 

All networks were optimized using Adam optimizer \cite{kingma2014adam} with an initial learning rate of $0.001$, batch size of $1$ and trained patches of size $128\times96\times64$ with stride between patches of size $32\times24\times16$ and a total of $100$ epochs within the training set. For (3) the model was first pretrained with $100$ epochs of $\langle \mathbf{I},\mathbf{J}\rangle$ pairs and afterwards fine-trained with $40$ epochs of $\langle \mathbf{I},\mathbf{L}\rangle$ pairs. In all experiments data augmentation (random deformation, random flipping, random rotation) was used.

\setlength{\tabcolsep}{3pt}
\begin{table*}[t]
  \begin{center}
    \caption{Segmentation accuracy for Y-Net-TL (pretrained with T1w/T1wce pairs and fine-tuned with T1w, T1wce and segmentations triplets). All segmentation results are for applying the model to non-contrast enhanced T1w data.}
    \label{tab:exp3_seg}
    \begin{tabular}{l | c c c c c c}
      Model & Dice Score & Jaccard Score & Mean Surface Distance & Recall & Precision & \textcolor{Black}{\makecell{Absolute\\VVR Difference}}\\
      \hline
      Y-Net-TL (5 T1w/T1wce) & 0.521 (0.09) & 0.357 (0.08) & 3.523 (1.10) & 0.435 (0.10) & 0.679 (0.12) & 1.843\\
      Y-Net-TL (10 T1w/T1wce) & 0.519 (0.10) & 0.357 (0.09) & 3.628 (1.35) & 0.446 (0.12) & 0.666 (0.14) & 1.900\\
      Y-Net-TL (20 T1w/T1wce) & 0.524 (0.09) & 0.359 (0.08) & 3.411 (1.13) & 0.447 (0.10) & 0.670 (0.14) & 1.840\\
      Y-Net-TL (30 T1w/T1wce) & 0.535 (0.10) & 0.371 (0.09) & 3.476 (1.11) & 0.465 (0.12) & 0.660 (0.12) & 1.660\\
    \end{tabular}
  \end{center}
\end{table*}

\subsection{Improving vessel segmentation with an auxiliary training target}\label{sec:exp1}

The multi-task Y-Net improves the overall Dice score of vessel segmentation on T1w data from $0.446$ to $0.506$ {\textcolor{Black}{(p = $0.09$)}}, and Jaccard Score from $0.294$ to $0.343$ {\textcolor{Black}{(p = $0.09$)}} when compared to a U-Net segmenting based on the same image data in experiment 1 (Table \ref{tab:exp1_seg}). Multi-task learning decreases mean surface distance from $4.615$ to $3.929$ {\textcolor{Black}{(p = $0.26$)}} and results in gains in recall and slight decrease in precision for vessel segmentation compared to a single-task U-Net. \textcolor{Black}{Frangi filter \cite{frangi1998multiscale} in combination with Otsu thresholding \cite{otsu1975threshold} resulted in a Dice score of $0.020$, Jaccard score of $0.010$ and a mean surface distance of $75.822$.}\textcolor{Black}{}

\subsection{\textcolor{Black}{Improving} vessel segmentation with non-annotated imaging data}\label{sec:exp2}

In experiment 2 we trained a Y-Net-mix model with additional examples without annotations (Fig.~\ref{fig:train}b). Adding $\langle \mathbf{I},\mathbf{J}\rangle$ pairs further improved segmentation accuracy compared with the Y-Net trained only on annotated cases (Table \ref{tab:exp2_seg}). With 30 additional examples the Dice score increased from $0.506$ to $0.539$ {\textcolor{Black}{(p < 0.01)}}, the Jaccard score from $0.343$ to $0.376$ {\textcolor{Black}{(p < 0.01)}} and the mean surface distance decreased from $3.929$ to $3.621$ {\textcolor{Black}{(p = 0.15)}}. Recall of Y-Net-mix is better than U-Net and Y-Net, precision is better for the U-Net. Further analysis in section \ref{sec:exp4} revealed that this increase was due to small diameter vessels. 
%
The multi-task loss function and automated search for optimal $\sigma_{S}$ and $\sigma_{T}$ is not stable when the number of training samples for each task is substantially different. Therefore, we fixed $\sigma_{S}$ and $\sigma_{T}$ to values learned in the first experiment.

\begin{figure}[H]
\centering
\subfigure[]{\includegraphics[width=\columnwidth]{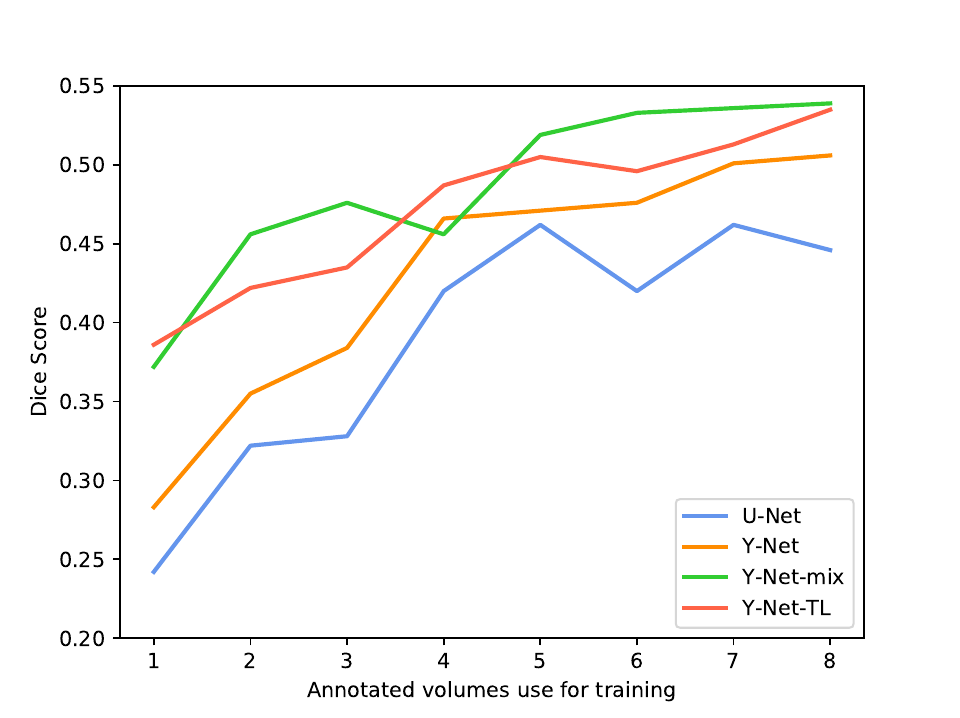}}
\subfigure[]{\includegraphics[width=\columnwidth]{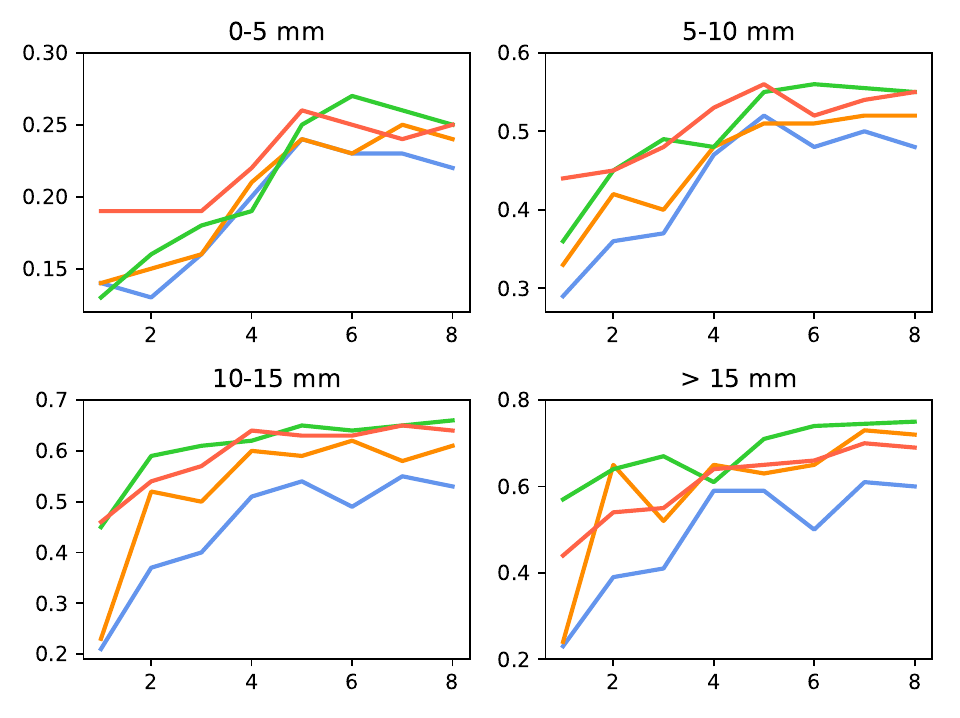}}
\caption{Segmentation accuracy (a) in relation to number of segmentations used for training for the whole vessel tree and (b) different vessel sizes between baseline and proposed methodologies. Multi-task learning with additional T1w/T1wce pairs benefits more if less annotations are used for training.} 
\label{fig:acc_num_seg}
\end{figure}

\subsection{Combining transfer learning and multi-task learning}\label{sec:exp3}

The third experiment evaluated an alternative approach of combining data with and without annotations. Here, the Y-Net-TL was pretrained with $5$ to $30$ $\langle \mathbf{I},\mathbf{J}\rangle$ pairs (Fig.~\ref{fig:train}c) for $100$ epochs. Then it was fine tuned (Fig.~\ref{fig:train}d) with $8$ $\langle \mathbf{I},\mathbf{J},\mathbf{L}\rangle$ triplets for $40$ epochs in a $3$-fold cross validation ($8$ training samples, $4$ test samples). Pretraining the Y-Net-TL with $30$ T1w/T1wce pairs improved the Dice score from $0.506$ to $0.535$ {\textcolor{Black}{(p < 0.01)}}, decreased the mean surface distance from $3.929$ to $3.476$ {\textcolor{Black}{(p = 0.06)}}, and improved recall over the Y-Net~(Table \ref{tab:exp3_seg}). The successive addition of T1w/T1wce training pairs for pretraining resulted in increasingly better segmentation accuracy (Dice). Overall accuracy is comparable to the strategy evaluated in section \ref{sec:exp2}.  

\setlength{\tabcolsep}{3pt}
\begin{table*}
  \begin{center}
    \caption{Dice, mean surface distance, recall and precision related to the vessel thickness between baseline U-Net, proposed Y-Net, Y-Net-mix trained with additional 30 T1w/T1wce training pairs without annotation and Y-Net-TL (Y-Net is pretrained with 30 T1w/T1wce pairs without annotations and fine tuned with T1w, T1wce and segmentation). Percentage values relate to affiliation of total vessel volume.}
    \label{tab:expx_seg_thick}
    \begin{tabular}{l l | c c c c c c}
      \multicolumn{2}{r|}{Vessel thickness:} & $0-5$ mm  & $5-10$ mm & $10-15$ mm & $>15$ mm \\
      & & ($34.489\%$) & ($31.835\%$) & ($16.229\%$) & ($17.447\%$) \\
      \hline
      \multirow{2}{*}{Dice} & U-Net & 0.22 (0.08)  & 0.48 (0.09) & 0.53 (0.21) &  0.60 (0.22) \\ 
      & Y-Net & 0.24 (0.09) & 0.52 (0.07) & 0.61 (0.12) & 0.72 (0.05) \\ 
      & Y-Net-mix & 0.25 (0.09) & 0.55 (0.08) & 0.66 (0.10) & 0.75 (0.06) \\ 
      & Y-Net-TL & 0.25 (0.08) & 0.55 (0.09) & 0.64 (0.13) & 0.69 (0.06) \\ \hline
      \multirow{2}{*}{Mean Surface Distance} & U-Net & 7.11 (3.30) & 4.4 (2.81) & inf & 5.44 (3.29)  \\ 
      & Y-Net & 6.41 (3.00)  & 3.18 (1.15) & 2.76 (1.66) & 3.15 (1.32) \\ 
      & Y-Net-mix & 6.07 (3.16) & 3.06 (1.91) & 2.60 (1.71) & 2.34 (1.06)  \\ 
      & Y-Net-TL & 5.56 (2.45) & 2.95 (1.20) & 3.06 (2.50) & 2.93 (1.29)\\ \hline      
      \multirow{2}{*}{Recall} & U-Net & 0.17 (0.09)  & 0.39 (0.12) & 0.44 (0.21) &  0.51 (0.25) \\ 
      & Y-Net & 0.20 (0.10) & 0.43 (0.09) & 0.53 (0.15) & 0.64 (0.12) \\ 
      & Y-Net-mix & 0.22 (0.11) & 0.49 (0.12) & 0.60 (0.12) & 0.70 (0.13) \\ 
      & Y-Net-TL & 0.21 (0.10) & 0.48 (0.12) & 0.59 (0.18) & 0.69 (0.10) \\ \hline
      \multirow{2}{*}{Precision} & U-Net & 0.41 (0.08)  & 0.69 (0.11) & 0.73 (0.24) &  0.88 (0.06) \\ 
      & Y-Net & 0.36 (0.07) & 0.69 (0.10) & 0.77 (0.06) & 0.86 (0.09) \\ 
      & Y-Net-mix & 0.37 (0.08) & 0.69 (0.14) & 0.76 (0.08) & 0.85 (0.10) \\ 
      & Y-Net-TL & 0.36 (0.06) & 0.67 (0.06) & 0.77 (0.09) & 0.84 (0.11) \\ \hline
    \end{tabular}
  \end{center}
\end{table*}

\subsection{Learning with few annotated examples}

We evaluated the benefit of multi-task learning if particularly few annotations are available.  We compared a U-Net, Y-Net (Sec.\,\ref{sec:exp1}), a Y-Net-mix (Sec.\,\ref{sec:exp2}), and Y-Net-TL (Sec.\,\ref{sec:exp3}) varying the number of annotated examples from $1$ to $8$ (Fig. \ref{fig:acc_num_seg}a). For a single annotated example the U-Net achieved a Dice of $0.242$, the Y-Net $0.283$ {\textcolor{Black}{(p < 0.01)}}, and the Y-Net-mix with additional 30 cases $0.372$ {\textcolor{Black}{(p < 0.01)}}. Pretraining the Y-Net-TL with $30$ T1w/T1wce pairs yielded $0.386$ {\textcolor{Black}{(p < 0.001)}}. When increasing the number of annotated volumes from $1$ and $8$ the gap between U-Net and Y-Net-mix decreased from $0.130$ to $0.093$. 

\begin{figure}
\centerline{\includegraphics[width=\columnwidth]{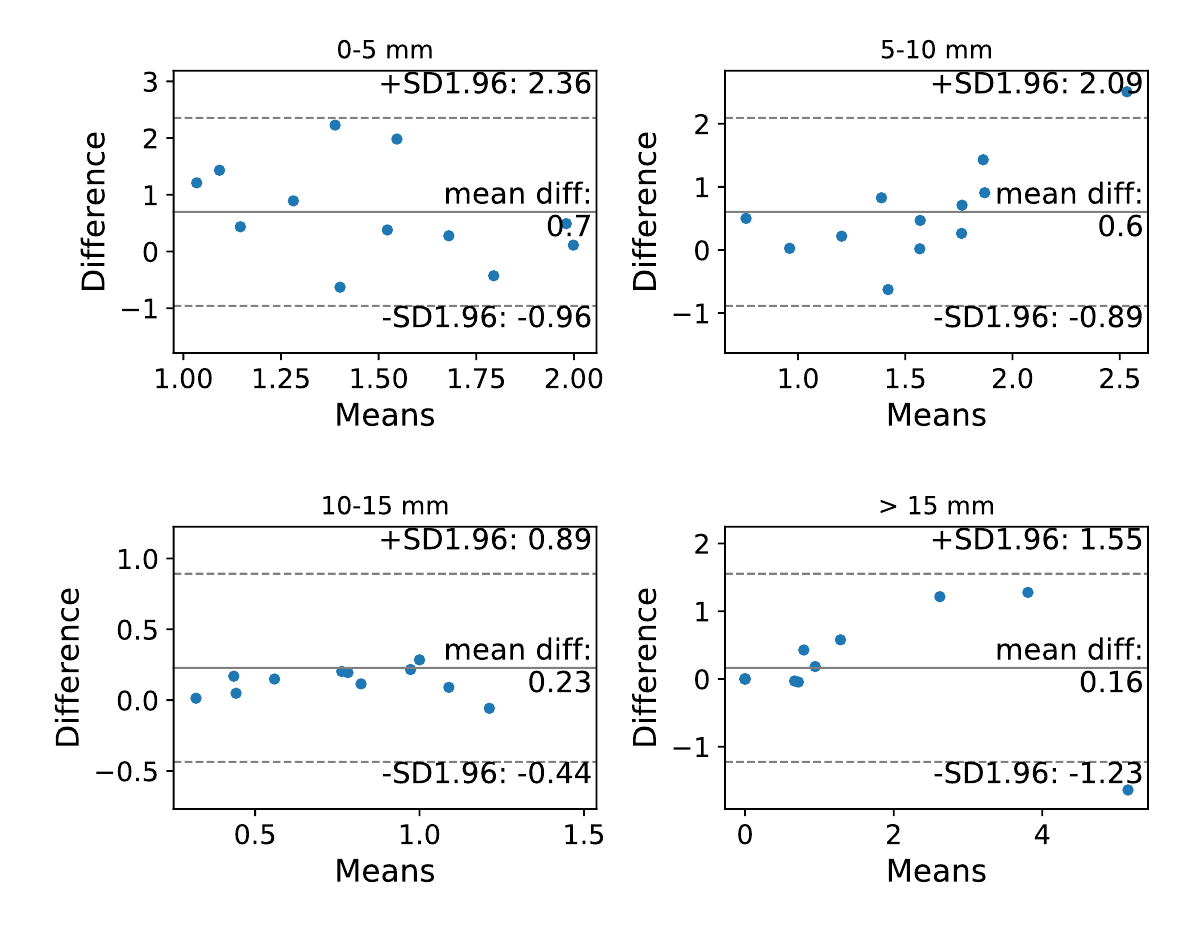}}
\caption{Bland-Altman plot for vessel to volume ratio between ground truth and Y-Net-mix + 30 T1w/T1wce pairs for the four investigated vessel sizes. With increasing vessel size the mean difference is decreased.} 
\label{fig:altman}
\end{figure}

\subsection{Segmentation accuracy for different vessel diameters}\label{sec:exp4}

\begin{figure*}[t]
\centerline{\includegraphics[width=\textwidth]{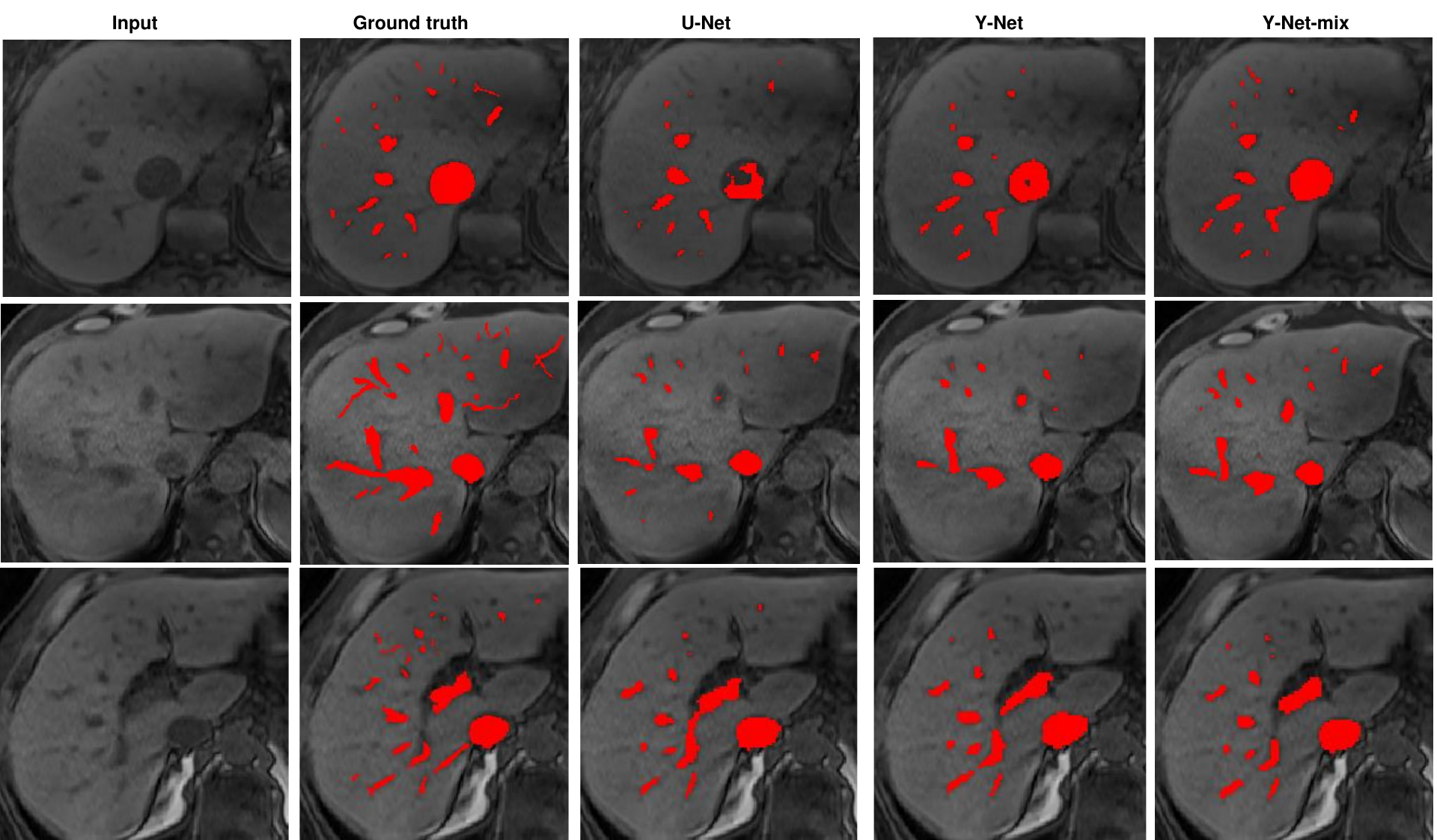}}
\caption{Qualitative example of segmentation results. T1w image, T1w image with ground truth vessel segmentation, and predictions within U-Net and the proposed Y-Net and the Y-Net-mix trained with additional 30 T1w/T1wce pairs. All networks are trained with the same annotations.} 
\label{fig:results}
\end{figure*}

\begin{table*}[t]
  \begin{center}
    \caption{Segmentation accuracy from hold-out testset between baseline U-Net, proposed Y-Net, Y-Net-mix trained with $30$ additional T1w/T1wce training pairs without annotation and Y-Net-TL (Y-Net is pretrained with $30$ T1w/T1wce pairs without annotations and fine tuned with T1w, T1wce and segmentation). All networks are trained with the same annotations.}
    \label{tab:exp5_seg}
    \begin{tabular}{l | c c c c c c }
      Model & Dice Score & Jaccard Score & Mean Surface Distance & Recall & Precision & \textcolor{Black}{Absolute VVR Difference}\\
      \hline
      U-Net & 0.474 (0.05) & 0.312 (0.04) & 3.256 (0.39) & 0.501 (0.11) & 0.488 (0.12) & 0.920 \\ 
      Y-Net & 0.498 (0.03) & 0.332 (0.03) & 3.222 (0.38) & 0.545 (0.07) & 0.489 (0.12) & 0.979\\ 
      Y-Net-mix & 0.504 (0.05) & 0.339 (0.05) & 3.109 (0.32) & 0.547 (0.14) & 0.514 (0.13) & 1.042 \\
      Y-Net-TL &  0.456 (0.07) & 0.299 (0.06) & 3.199 (0.39) & 0.502 (0.18) & 0.473 (0.09) & 1.076\\
    \end{tabular}
  \end{center}
\end{table*}

All architectures exhibited similar behavior with accuracy increasing with higher vessel diameter. Y-Net-mix performed best across all diameters (Table \ref{tab:expx_seg_thick}). Compared to the U-Net, Y-Net-mix improved the Dice score in bigger vessel ($>15$ mm: $+0.15$, $10-15$ mm: $+0.13$) more as in smaller vessels ($5-10$mm: $+0.07$, $0-5$mm: $+0.03$) with gains in recall and slight decrease in precision. Bland-Altman plot (Fig. \ref{fig:altman}) showed decreased vessel to volume ratio mean difference with increasing vessel size between ground truth and the Y-Net + 30 T1w/T1wce pairs. If less than $8$ annotations are used for training (Fig. \ref{fig:acc_num_seg}b), results are similar as with $8$ annotations, except if only $1$ segmentation is used for training then the Y-Net-TL showed best Dice results in segmenting small vessels with diameter between $0-5$ ($+0.05$) and $5-10$ mm ($+0.15$) compared to the baseline U-Net.
These results suggest that the proposed multi-task learning is particularly relevant in cases where few annotated training examples are available, but holds the advantage to a lesser degree even if more annotated training examples are used.

\subsection{Accuracy on a hold-out testset}\label{sec:exp5}

In the last experiment we evaluated the segmentation accuracy from U-Net, Y-Net and the best Y-net-mix and Y-net-TL models (based on mean surface distance) on a hold-out test set. This data set contained a total of $3$ chronic liver diseased volumes. For this experiment we trained every model with all $12$ annotations from the cross-validation runs and applied to the hold-out set (Table \ref{tab:exp5_seg}). Results are consistent with the cross-validation data. Compared to the U-net, the Y-Net increased the Dice score from $0.474$ to $0.498$, and the Y-Net-mix with additional $30$ T1w/T1wce pairs further increased the accuracy from $0.474$ to $0.504$ compared to the baseline U-Net. The Y-Net-TL ($30$ pre-training examples) performs worse than the U-net. Further, the Y-net-mix decreased the mean surface distance from $3.256$ to $3.109$, improved recall from $0.501$ to $0.547$ and increased precision from $0.488$ to $0.514$ compared to the U-Net. {\textcolor{Black}{Due to the small size of the hold-out testset the p-values are not significant.}}

\setlength{\tabcolsep}{3pt}
\begin{table*}
  \begin{center}
    \caption{Dice, mean surface distance, recall and precision related to the vessel thickness from hold-out testset between the baseline U-Net, the proposed Y-Net and the Y-Net-mix trained with additional 30 T1w/T1wce pairs and the Y-Net-TL (Y-Net is pretrained with 30 
T1w/T1wce pairs without annotations and fine tuned with T1w, T1wce and segmentation). Percentage values relate to affiliation of total vessel volume.}
    \label{tab:exp5_seg_thick}
    \begin{tabular}{ l l | c c c c c c}
      \multicolumn{2}{r|}{Vessel thickness:} & $0-5$ mm  & $5-10$ mm & $10-15$ mm & $>15$ mm \\
      & & ($32.791\%$) & ($30.567\%$) & ($19.590\%$) & ($17.051\%$) \\
      \hline
      \multirow{2}{*}{Dice} & U-Net & 0.28 (0.02) & 0.52 (0.01) & 0.53 (0.10) & 0.68 (0.09) \\ 
      & Y-Net & 0.29 (0.02) & 0.54 (0.01) & 0.65 (0.04) & 0.75 (0.01)  \\ 
      & Y-Net-mix & 0.30 (0.02) & 0.56 (0.01) & 0.66 (0.07) & 0.69 (0.14)  \\ 
      & Y-Net-TL & 0.30 (0.01) & 0.56 (0.02) & 0.52 (0.19) & 0.59 (0.25) \\ \hline
      \multirow{2}{*}{Mean Surface Distance} & U-Net & 4.71 (0.69) & 3.60 (0.94) & 2.81 (0.73) & 2.84 (0.01)   \\ 
      & Y-Net & 4.52 (0.63) & 3.53 (0.72) & 2.37 (0.62) & 2.60 (1.40)   \\ 
      & Y-Net-mix & 4.54 (0.98) & 2.79 (0.62) & 2.35 (0.59) & 1.92 (0.71)   \\ 
      & Y-Net-TL & 4.22 (0.52) & 2.81 (0.19) & 3.17 (1.30) & 2.34 (1.17) \\ \hline      
      \multirow{2}{*}{Recall} & U-Net & 0.37 (0.10) & 0.52 (0.07) & 0.47 (0.13) & 0.56 (0.13) \\ 
      & Y-Net & 0.39 (0.12) & 0.54 (0.08) & 0.69 (0.09) & 0.64 (0.01) \\ 
      & Y-Net-mix & 0.38 (0.13) & 0.58 (0.10) & 0.68 (0.14) & 0.59 (0.20)   \\ 
      & Y-Net-TL &  0.39 (0.11) & 0.57 (0.10) & 0.50 (0.25) & 0.51 (0.30) \\ \hline
      \multirow{2}{*}{Precision} & U-Net & 0.28 (0.12) & 0.55 (0.11) & 0.63 (0.06) & 0.92 (0.05)  \\ 
      & Y-Net & 0.28 (0.11) & 0.56 (0.10) & 0.62 (0.04) & 0.91 (0.02)   \\ 
      & Y-Net-mix & 0.31 (0.14) & 0.59 (0.11) & 0.66 (0.07) & 0.93 (0.06)   \\ 
      & Y-Net-TL & 0.28 (0.08) & 0.58 (0.08) & 0.64 (0.03) & 0.91 (0.05)  \\ \hline
    \end{tabular}
  \end{center}
\end{table*}

Fig. \ref{fig:results} shows qualitative examples of segmentation results from three datasets, where improvement from Y-Net over U-Net and Y-Net trained with additionally 30 T1w/T1wce pairs over Y-Net is visible. All four architectures exhibit similar behavior with accuracy increasing with increasing vessel diameter (Table \ref{tab:exp5_seg_thick}). Across all diameters, the Y-Net and Y-Net-mix yielded consistent improvement in Dice, mean surface distance, recall and precision. The pre-trained and fine-tuned Y-Net-TL showed in one dataset poor performance in the $>15$ mm diameter group, which led to the overall lower accuracy compared to the U-Net. Segmentation accuracy is still related to the vessel thickness, suggesting that smaller vessels still require more annotated training data. This is also visible in Fig. \ref{fig:results}, where all three models lack in performance in segmenting smaller vessels in the three visualized datasets. Overall the Y-Net-mix trained with additional data outperformed the U-Net model.

\subsection{Validation on a brain tumor segmentation task}\label{sec:exp6}

\setlength{\tabcolsep}{3pt}
\begin{table*}[t]
  \begin{center}
    \caption{Segmentation accuracy for non-enhancing and enhancing tumor labels of the brain tumor segmentation dataset (and standard deviation between the datasets) with $8$ datasets used for training between the proposed Y-Net-mix trained with $30$ additional T2w/T1wce training pairs and the baseline U-Net}
    \label{tab:validation_seg}
    \begin{tabular}{l | c c c c }
    & \multicolumn{4}{c}{Non enhancing tumor} \\
      Model & Dice Score & Mean Surface Distance & Recall & Precision\\ \hline
      U-Net &  0.154 (0.18) & 11.393 (8.54) & 0.122 (0.16) & 0.379 (0.36) \\ 
      Y-Net-mix & 0.223 (0.22) & 10.192 (13.22) & 0.205 (0.23) & 0.432 (0.35) \\
      \\
    & \multicolumn{4}{c}{Enhancing tumor} \\
      Model & Dice Score & Mean Surface Distance & Recall & Precision\\ \hline
      U-Net &  0.144 (0.18) & 11.717 (9.76) & 0.132 (0.19) & 0.335 (0.30) \\ 
      Y-Net-mix & 0.214 (0.20) & 8.742 (7.85) & 0.205 (0.21) & 0.373 (0.30) \\      
    \end{tabular}
  \end{center}
\end{table*}

We evaluated the proposed approach on the publicly available brain tumor segmentation dataset from the medical segmentation decathlon \cite{simpson2019large}. The data set is challenging with scans from 19 institutions and field strengths ranging from $1$T to $3$T, with typically not all scanner types present in the training data. We used the T2-weighted (T2w) sequence as input and the T1-weighted contrast enhanced sequence as auxiliary data during training. We segmented non-enhancing- and enhancing tumor, labels for which contrast enhanced T1 MRI is highly informative. We performed the experiment $3$ times and sampled each time randomly $68$ examples ($8$ for training, $30$ for evaluation and $30$ as auxiliary examples) and compared segmentation accuracy between U-Net and Y-Net-mix with additional $30$ T2w/T1wce examples. 

\begin{figure}[t]
\centerline{\includegraphics[width=\columnwidth]{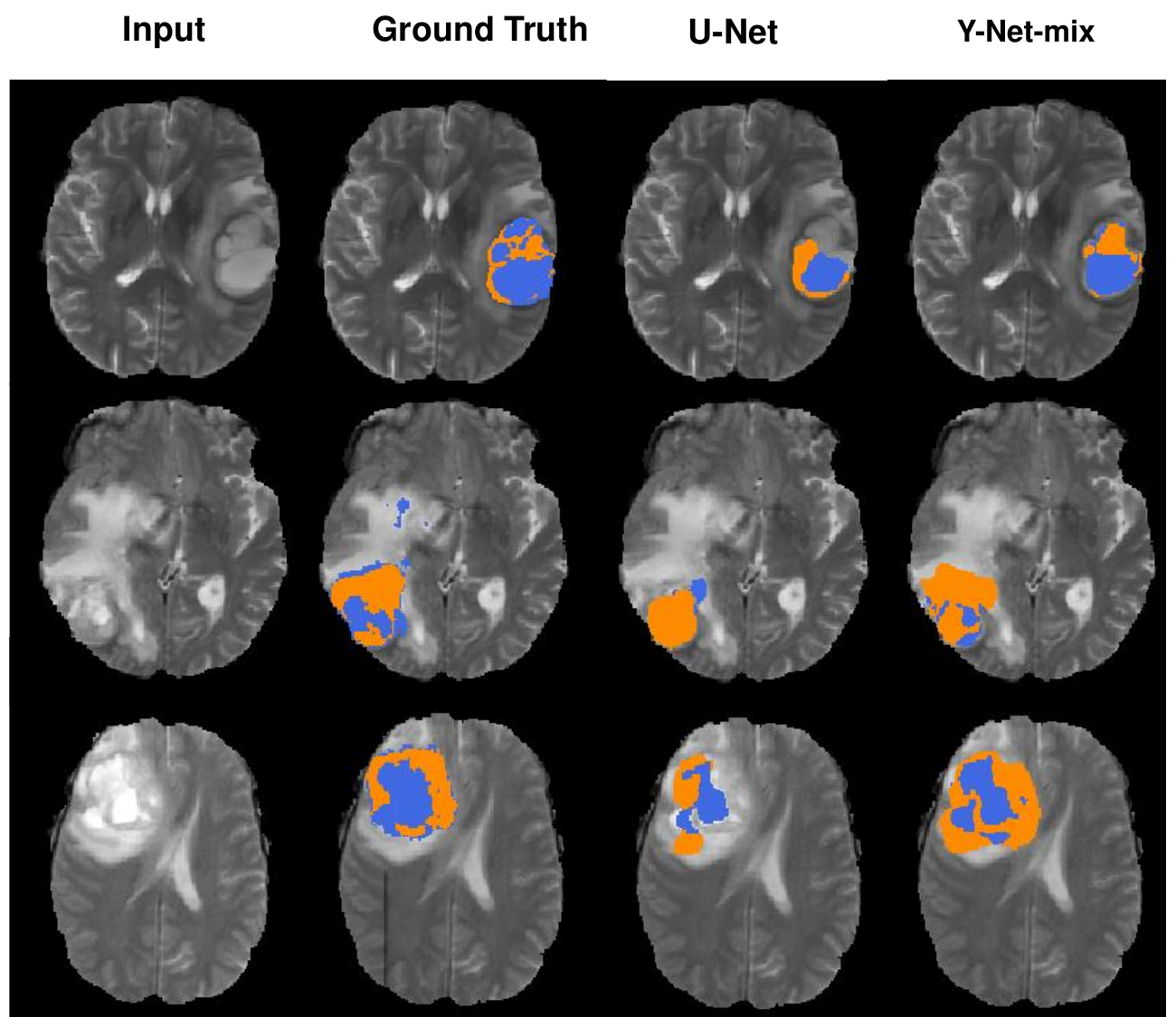}}
\caption{Qualitative example of segmentation results on the validation dataset. T2w image, T2w image with ground truth brain tumor segmentation, and predictions in within U-Net and the proposed Y-Net-mix trained with additional 30 T2w/T1wce pairs. The blue label denotes the non-enhancing tumor and the orange label the enhancing tumor. All networks are trained with the same annotations.} 
\label{fig:results_val_two_labels}
\end{figure}

\begin{figure*}[t]
\centerline{\includegraphics[width=\textwidth]{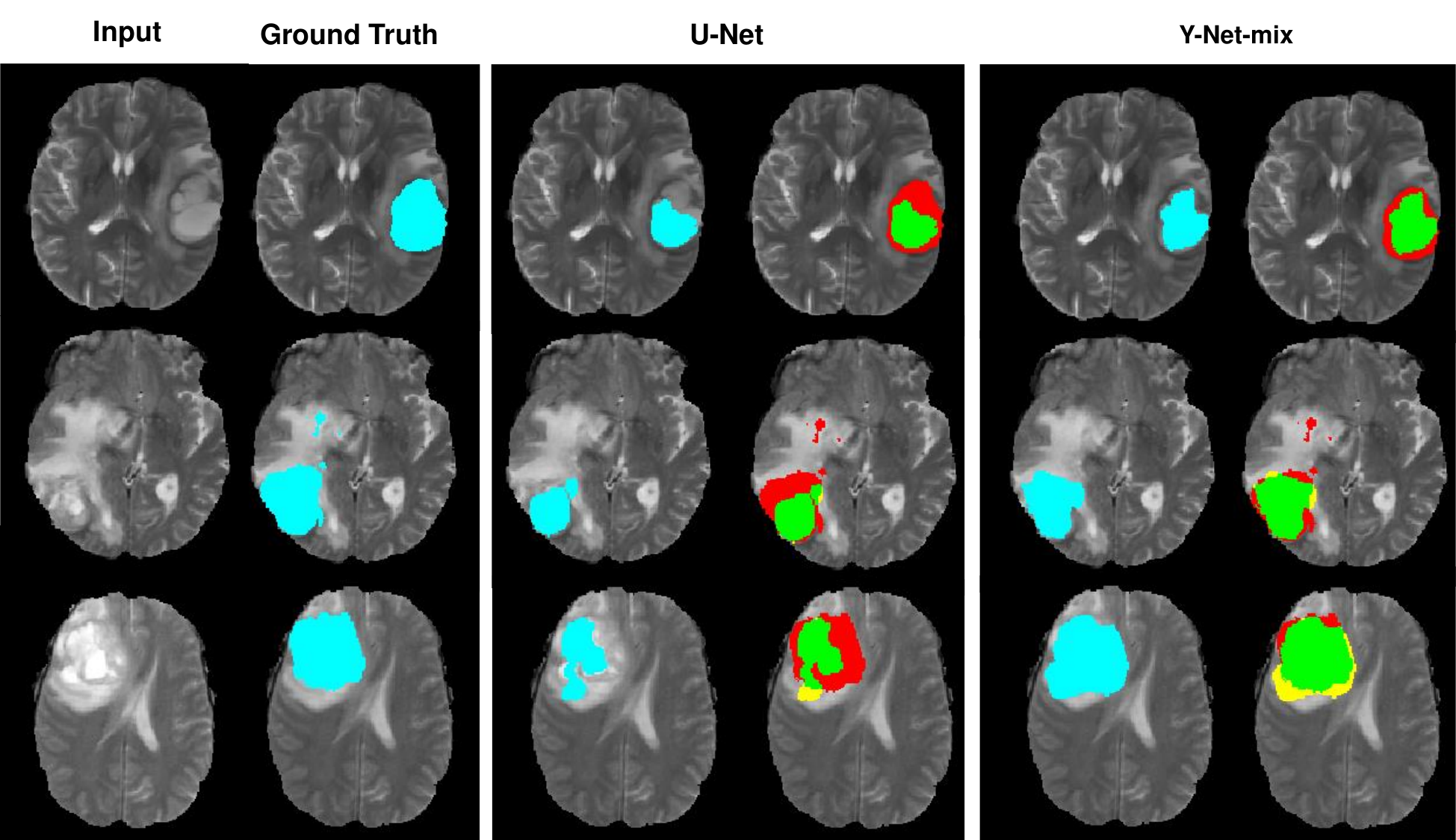}}
\caption{Qualitative example of segmentation results on the validation dataset. T2w image, T2w image with ground truth brain tumor segmentation, and predictions in within U-Net and the proposed Y-Net-mix trained with additional 30 T2w/T1wce pairs. Red denotes the false negative, green the true positive and yellow the false positive voxels. All networks are trained with the same annotations.} 
\label{fig:results_val}
\end{figure*}

When compared to a U-Net, a multi-task Y-Net-mix trained with additional $30$ T2w/T1wce cases increased the non-enhancing tumor Dice from $0.154$ to $0.223$ {\textcolor{Black}{(p < 0.001)}}, recall from $0.122$ to $0.205$ and precision from $0.379$ to $0.432$ (Table \ref{tab:validation_seg}). For the enhancing tumor label the Dice score increases from $0.144$ to $0.214$ {\textcolor{Black}{(p < 0.01)}}, recall from $0.132$ to $0.205$ and precision from $0.335$ to $0.373$. For the evaluation of the mean surface distance, cases without complete ground truth annotation were excluded ($15$ from $90$ datasets for the non-enhancing tumor label and $19$ from $90$ datasets for the enhancing tumor label). The mean surface distance decreases from $11.393$ (U-net) to $10.192$ (Y-net-mix) {\textcolor{Black}{(p = 0.39)}} for the non-enhancing tumor and decreases from $11.717$ (U-net) to $8.742$ (Y-net-mix) {\textcolor{Black}{(p < 0.01)}} for the enhancing tumor. Qualitative results are shown in Figure \ref{fig:results_val_two_labels} and \ref{fig:results_val}. 

\section{Discussion}\label{sec:Conclusion}

We propose a 3D multi-task learning approach for medical image segmentation. An auxiliary modality available only during training improves the segmentation accuracy of a Y-Net model when applied to new data, even if it is not available during test time. We show that providing such privileged informative data during training has benefits especially, if annotated training examples are scarce. In our experiments this improves segmentation of vessels and brain tumors in MRI without contrast enhancement when contrast enhanced MRI is used as auxiliary image type during training. 

Results demonstrate that an auxiliary training modality improves segmentation accuracy, even if no annotation is available for a large part of the training data. Reproducing the benefits of multi-task learning on the segmentation of liver vessels and brain tumors in MRI data suggests that the strategy is widely applicable. CNN style multi-task learning models are able to benefit from the similarity of annotations and informative modalities. Instead of limiting model training to few widely available data types, the proposed technique enables the inclusion of highly informative imaging (contrast enhanced MRI) during training and transfers benefits to the model inference on limited data (MRI without contrast enhancement) during test time. 

This is relevant for models applicable across a diverse set of contexts or in clinical studies for which minimal feasible protocols are implemented. 
Results suggest that multi-task learning is able to exploit the relationship between informative image modalities and the target labels, benefiting from additional variability that is contributed by having more training data.  

This work has limitations. Despite a large number of annotated MRI slices, the number of annotated images from different patients is small. However, the consistency of the cross-validation results and results on the hold-out data suggests their validity. Another limitation is the multi-task loss function, which works well when the model is trained on both tasks simultaneously, but lacks in performance when training imbalance of the two tasks is different (like in section \ref{sec:exp2}).

In conclusion, the findings are relevant for segmentation tasks, where annotation is costly, and where dedicated MR sequences with good contrast for the structure of interest are available during training, but not widely available during the application of models.

\textcolor{Black}{Liver vessel segmentation can be used for planning liver resection of primary liver tumour, such as hepatocellular carcinoma complicating CLD, or rarely metastases in the cirrhotic liver \cite{alirr2023hepatic}. \cite{beichel2004liver} proposed a further application, i.e., using the portal veins to divide the liver into segments. Vessel segmentation can also be used to assess the vasculature of the living donor before liver transplantation for HCC or liver failure \cite{alirr2018automated}. In the broader context, portal vein segmentation may allow us to study vascular features in CLD patients with known portal hypertension.  Indeed, \cite{sun2020vascular} found that the branch angle, as well as deviation from Murray’s cube, which defines the relationship between the hepatic artery and portal vein, were significantly correlated with the portal hypertension.  Thus, it is conceivable that by looking at the 3D microvascular pattern of the portal venous system, we may be able to identify features that are associated with portal hypertension in humans. It is known that fibrosis causes microvascular changes in the liver that increase resistance to blood flow, directly increasing portal pressures \cite{sun2020vascular}. Thus, by identifying markers of portal hypertension, we might one day be able to non-invasively diagnose portal hypertension rather than placing a catheter in the portal vein to measure the hepatic venous pressure gradient (HVPG).  The lack of reliance on this highly invasive procedure, which is observer-dependent and not widely available, would be a great leap forward.}

\section*{Acknowledgement}
This work has been partially funded by the Vienna Science and Technology Fund (WWTF) [10.47379/LS20065], Austrian Science Fund FWF [P 35189], the European Union’s Horizon Europe research and innovation programme under grant agreement No. 101080302 AI-POD, No. 101136299 ARTEMIS and Novartis Pharmaceuticals Corporation.

\textcolor{Black}{\section*{Data and code availability statement}
The code of the multi-task learning framework is available at \url{https://github.com/cirmuw/mtl-ynet} and the baseline 3D-UNet is adopted from \cite{wolny2020accurate}.}

\bibliographystyle{IEEEbib}
\bibliography{cas-refs}

\begin{thebibliography}{10}

\bibitem{khawaja2015revisiting}
Aurang~Z Khawaja, Deirdre~B Cassidy, Julien Al~Shakarchi, Damian~G McGrogan, Nicholas~G Inston, and Robert~G Jones,
\newblock ``Revisiting the risks of {MRI} with gadolinium based contrast agents—review of literature and guidelines,''
\newblock {\em Insights into imaging}, vol. 6, no. 5, pp. 553--558, 2015.

\bibitem{scaglioni2011ash}
F~Scaglioni, S~Ciccia, M~Marino, G~Bedogni, and S~Bellentani,
\newblock ``Ash and nash,''
\newblock {\em Digestive Diseases}, vol. 29, no. 2, pp. 202--210, 2011.

\bibitem{lakshman2015synergy}
Raj Lakshman, Ruchi Shah, Karina Reyes-Gordillo, and Ravi Varatharajalu,
\newblock ``Synergy between nafld and afld and potential biomarkers,''
\newblock {\em Clinics and research in hepatology and gastroenterology}, vol. 39, pp. S29--S34, 2015.

\bibitem{alirr2023hepatic}
Omar~Ibrahim Alirr and Ashrani Aizzuddin~Abd Rahni,
\newblock ``Hepatic vessels segmentation using deep learning and preprocessing enhancement,''
\newblock {\em Journal of applied clinical medical physics}, vol. 24, no. 5, pp. e13966, 2023.

\bibitem{alirr2018automated}
Omar~Ibrahim Alirr, Ashrani Aizzuddin~Abd Rahni, and Ehsan Golkar,
\newblock ``An automated liver tumour segmentation from abdominal ct scans for hepatic surgical planning,''
\newblock {\em International journal of computer assisted radiology and surgery}, vol. 13, pp. 1169--1176, 2018.

\bibitem{bastati2020does}
Nina Bastati, Lucian Beer, Mattias Mandorfer, Sarah Poetter-Lang, Dietmar Tamandl, Yesim Bican, Michael~Christoph Elmer, Henrik Einspieler, Georg Semmler, Benedikt Simbrunner, et~al.,
\newblock ``Does the functional liver imaging score derived from gadoxetic acid--enhanced {MRI} predict outcomes in chronic liver disease?,''
\newblock {\em Radiology}, vol. 294, no. 1, pp. 98--107, 2020.

\bibitem{ciecholewski2021computational}
Marcin Ciecholewski and Micha{\l} Kassja{\'n}ski,
\newblock ``Computational methods for liver vessel segmentation in medical imaging: A review,''
\newblock {\em Sensors}, vol. 21, no. 6, pp. 2027, 2021.

\bibitem{frangi1998multiscale}
Alejandro~F Frangi, Wiro~J Niessen, Koen~L Vincken, and Max~A Viergever,
\newblock ``Multiscale vessel enhancement filtering,''
\newblock in {\em International conference on medical image computing and computer-assisted intervention}. Springer, 1998, pp. 130--137.

\bibitem{sato1998three}
Yoshinobu Sato, Shin Nakajima, Nobuyuki Shiraga, Hideki Atsumi, Shigeyuki Yoshida, Thomas Koller, Guido Gerig, and Ron Kikinis,
\newblock ``Three-dimensional multi-scale line filter for segmentation and visualization of curvilinear structures in medical images,''
\newblock {\em Medical image analysis}, vol. 2, no. 2, pp. 143--168, 1998.

\bibitem{meijering2004design}
Erik Meijering, M~Jacob, J-CF Sarria, Pl~Steiner, H~Hirling, and M~Unser,
\newblock ``Design and validation of a tool for neurite tracing and analysis in fluorescence microscopy images,''
\newblock {\em Cytometry Part A: the journal of the International Society for Analytical Cytology}, vol. 58, no. 2, pp. 167--176, 2004.

\bibitem{soares2006retinal}
Jo{\~a}o~VB Soares, Jorge~JG Leandro, Roberto~M Cesar, Herbert~F Jelinek, and Michael~J Cree,
\newblock ``Retinal vessel segmentation using the 2-d gabor wavelet and supervised classification,''
\newblock {\em IEEE Transactions on medical Imaging}, vol. 25, no. 9, pp. 1214--1222, 2006.

\bibitem{lu2017hepatic}
Siyu Lu, Hui Huang, Ping Liang, Gang Chen, and Liang Xiao,
\newblock ``Hepatic vessel segmentation using variational level set combined with non-local robust statistics,''
\newblock {\em Magnetic resonance imaging}, vol. 36, pp. 180--186, 2017.

\bibitem{zeng2018automatic}
Ye-zhan Zeng, Sheng-hui Liao, Ping Tang, Yu-qian Zhao, Miao Liao, Yan Chen, and Yi-xiong Liang,
\newblock ``Automatic liver vessel segmentation using 3d region growing and hybrid active contour model,''
\newblock {\em Computers in biology and medicine}, vol. 97, pp. 63--73, 2018.

\bibitem{chung2018accurate}
Minyoung Chung, Jeongjin Lee, Jin~Wook Chung, and Yeong-Gil Shin,
\newblock ``Accurate liver vessel segmentation via active contour model with dense vessel candidates,''
\newblock {\em Computer methods and programs in biomedicine}, vol. 166, pp. 61--75, 2018.

\bibitem{lebre2019automatic}
Marie-Ange Lebre, Antoine Vacavant, Manuel Grand-Brochier, Hugo Rositi, Armand Abergel, Pascal Chabrot, and Benoit Magnin,
\newblock ``Automatic segmentation methods for liver and hepatic vessels from ct and mri volumes, applied to the couinaud scheme,''
\newblock {\em Computers in biology and medicine}, vol. 110, pp. 42--51, 2019.

\bibitem{guo2020novel}
Xiaoyu Guo, Ruoxiu Xiao, Tao Zhang, Cheng Chen, Jiayu Wang, and Zhiliang Wang,
\newblock ``A novel method to model hepatic vascular network using vessel segmentation, thinning, and completion,''
\newblock {\em Medical \& biological engineering \& computing}, pp. 1--16, 2020.

\bibitem{ibragimov2017combining}
Bulat Ibragimov, Diego Toesca, Daniel Chang, Albert Koong, and Lei Xing,
\newblock ``Combining deep learning with anatomical analysis for segmentation of the portal vein for liver sbrt planning,''
\newblock {\em Physics in Medicine \& Biology}, vol. 62, no. 23, pp. 8943, 2017.

\bibitem{huang2018robust}
Qing Huang, Jinfeng Sun, Hui Ding, Xiaodong Wang, and Guangzhi Wang,
\newblock ``Robust liver vessel extraction using 3d u-net with variant dice loss function,''
\newblock {\em Computers in biology and medicine}, vol. 101, pp. 153--162, 2018.

\bibitem{mishra2018segmentation}
Deepak Mishra, Santanu Chaudhury, Mukul Sarkar, Sidharth Manohar, and Arvinder~Singh Soin,
\newblock ``Segmentation of vascular regions in ultrasound images: A deep learning approach,''
\newblock in {\em 2018 IEEE International Symposium on Circuits and Systems (ISCAS)}. IEEE, 2018, pp. 1--5.

\bibitem{kitrungrotsakul2019vesselnet}
Titinunt Kitrungrotsakul, Xian-Hua Han, Yutaro Iwamoto, Lanfen Lin, Amir~Hossein Foruzan, Wei Xiong, and Yen-Wei Chen,
\newblock ``Vesselnet: A deep convolutional neural network with multi pathways for robust hepatic vessel segmentation,''
\newblock {\em Computerized Medical Imaging and Graphics}, vol. 75, pp. 74--83, 2019.

\bibitem{keshwani2020topnet}
Deepak Keshwani, Yoshiro Kitamura, Satoshi Ihara, Satoshi Iizuka, and Edgar Simo-Serra,
\newblock ``Topnet: Topology preserving metric learning for vessel tree reconstruction and labelling,''
\newblock in {\em International Conference on Medical Image Computing and Computer-Assisted Intervention}. Springer, 2020, pp. 14--23.

\bibitem{thomson2020mr}
Bart~R Thomson, Jasper~N Smit, Oleksandra~V Ivashchenko, Niels~FM Kok, Koert~FD Kuhlmann, Theo~JM Ruers, and Matteo Fusaglia,
\newblock ``Mr-to-us registration using multiclass segmentation of hepatic vasculature with a reduced 3d u-net,''
\newblock in {\em International Conference on Medical Image Computing and Computer-Assisted Intervention}. Springer, 2020, pp. 275--284.

\bibitem{yan2020attention}
Qingsen Yan, Bo~Wang, Wei Zhang, Chuan Luo, Wei Xu, Zhengqing Xu, Yanning Zhang, Qinfeng Shi, Liang Zhang, and Zheng You,
\newblock ``An attention-guided deep neural network with multi-scale feature fusion for liver vessel segmentation,''
\newblock {\em IEEE Journal of Biomedical and Health Informatics}, 2020.

\bibitem{xu2020training}
Minfeng Xu, Yu~Wang, Ying Chi, and Xiansheng Hua,
\newblock ``Training liver vessel segmentation deep neural networks on noisy labels from contrast ct imaging,''
\newblock in {\em 2020 IEEE 17th International Symposium on Biomedical Imaging (ISBI)}. IEEE, 2020, pp. 1552--1555.

\bibitem{marcan2014segmentation}
Marija Marcan, Denis Pavliha, Maja~Marolt Music, Igor Fuckan, Ratko Magjarevic, and Damijan Miklavcic,
\newblock ``Segmentation of hepatic vessels from {MRI} images for planning of electroporation-based treatments in the liver,''
\newblock {\em Radiology and oncology}, vol. 48, no. 3, pp. 267--281, 2014.

\bibitem{goceri2017vessel}
Evgin Goceri, Zarine~K Shah, and Metin~N Gurcan,
\newblock ``Vessel segmentation from abdominal magnetic resonance images: adaptive and reconstructive approach,''
\newblock {\em International journal for numerical methods in biomedical engineering}, vol. 33, no. 4, pp. e2811, 2017.

\bibitem{cciccek20163d}
{\"O}zg{\"u}n {\c{C}}i{\c{c}}ek, Ahmed Abdulkadir, Soeren~S Lienkamp, Thomas Brox, and Olaf Ronneberger,
\newblock ``3d u-net: learning dense volumetric segmentation from sparse annotation,''
\newblock in {\em International conference on medical image computing and computer-assisted intervention}. Springer, 2016, pp. 424--432.

\bibitem{caruana1997multitask}
Rich Caruana,
\newblock ``Multitask learning,''
\newblock {\em Machine learning}, vol. 28, no. 1, pp. 41--75, 1997.

\bibitem{zhang2021survey}
Yu~Zhang and Qiang Yang,
\newblock ``A survey on multi-task learning,''
\newblock {\em IEEE Transactions on Knowledge and Data Engineering}, 2021.

\bibitem{standley2019tasks}
Trevor Standley, Amir~R Zamir, Dawn Chen, Leonidas Guibas, Jitendra Malik, and Silvio Savarese,
\newblock ``Which tasks should be learned together in multi-task learning?,''
\newblock {\em arXiv preprint arXiv:1905.07553}, 2019.

\bibitem{weninger2019multi}
Leon Weninger, Qianyu Liu, and Dorit Merhof,
\newblock ``Multi-task learning for brain tumor segmentation,''
\newblock in {\em International MICCAI Brainlesion Workshop}. Springer, 2019, pp. 327--337.

\bibitem{amyar2020multi}
Amine Amyar, Romain Modzelewski, Hua Li, and Su~Ruan,
\newblock ``Multi-task deep learning based {CT} imaging analysis for covid-19 pneumonia: Classification and segmentation,''
\newblock {\em Computers in Biology and Medicine}, p. 104037, 2020.

\bibitem{zhang2017survey}
Yu~Zhang and Qiang Yang,
\newblock ``A survey on multi-task learning,''
\newblock {\em arXiv preprint arXiv:1707.08114}, 2017.

\bibitem{misra2016cross}
Ishan Misra, Abhinav Shrivastava, Abhinav Gupta, and Martial Hebert,
\newblock ``Cross-stitch networks for multi-task learning,''
\newblock in {\em Proceedings of the IEEE Conference on Computer Vision and Pattern Recognition}, 2016, pp. 3994--4003.

\bibitem{vandenhende2020revisiting}
Simon Vandenhende, Stamatios Georgoulis, Marc Proesmans, Dengxin Dai, and Luc Van~Gool,
\newblock ``Revisiting multi-task learning in the deep learning era,''
\newblock {\em arXiv preprint arXiv:2004.13379}, 2020.

\bibitem{kendall2018multi}
Alex Kendall, Yarin Gal, and Roberto Cipolla,
\newblock ``Multi-task learning using uncertainty to weigh losses for scene geometry and semantics,''
\newblock in {\em Proceedings of the IEEE conference on computer vision and pattern recognition}, 2018, pp. 7482--7491.

\bibitem{chen2018gradnorm}
Zhao Chen, Vijay Badrinarayanan, Chen-Yu Lee, and Andrew Rabinovich,
\newblock ``Gradnorm: Gradient normalization for adaptive loss balancing in deep multitask networks,''
\newblock in {\em International Conference on Machine Learning}. PMLR, 2018, pp. 794--803.

\bibitem{guo2018dynamic}
Michelle Guo, Albert Haque, De-An Huang, Serena Yeung, and Li~Fei-Fei,
\newblock ``Dynamic task prioritization for multitask learning,''
\newblock in {\em Proceedings of the European Conference on Computer Vision (ECCV)}, 2018, pp. 270--287.

\bibitem{liu2019end}
Shikun Liu, Edward Johns, and Andrew~J Davison,
\newblock ``End-to-end multi-task learning with attention,''
\newblock in {\em Proceedings of the IEEE Conference on Computer Vision and Pattern Recognition}, 2019, pp. 1871--1880.

\bibitem{li2020joint}
Lei Li, Xin Weng, Julia~A Schnabel, and Xiahai Zhuang,
\newblock ``Joint left atrial segmentation and scar quantification based on a dnn with spatial encoding and shape attention,''
\newblock in {\em International Conference on Medical Image Computing and Computer-Assisted Intervention}. Springer, 2020, pp. 118--127.

\bibitem{wang2019net}
Kaiqiang Wang, Jiazhen Dou, Qian Kemao, Jianglei Di, and Jianlin Zhao,
\newblock ``Y-net: a one-to-two deep learning framework for digital holographic reconstruction,''
\newblock {\em Optics Letters}, vol. 44, no. 19, pp. 4765--4768, 2019.

\bibitem{gao2019nddr}
Yuan Gao, Jiayi Ma, Mingbo Zhao, Wei Liu, and Alan~L Yuille,
\newblock ``Nddr-cnn: Layerwise feature fusing in multi-task cnns by neural discriminative dimensionality reduction,''
\newblock in {\em Proceedings of the IEEE/CVF Conference on Computer Vision and Pattern Recognition}, 2019, pp. 3205--3214.

\bibitem{wu2018group}
Yuxin Wu and Kaiming He,
\newblock ``Group normalization,''
\newblock in {\em Proceedings of the European conference on computer vision (ECCV)}, 2018, pp. 3--19.

\bibitem{lee1994building}
Ta-Chih Lee, Rangasami~L Kashyap, and Chong-Nam Chu,
\newblock ``Building skeleton models via 3-d medial surface axis thinning algorithms,''
\newblock {\em CVGIP: Graphical Models and Image Processing}, vol. 56, no. 6, pp. 462--478, 1994.

\bibitem{kingma2014adam}
Diederik~P Kingma and Jimmy Ba,
\newblock ``Adam: A method for stochastic optimization,''
\newblock {\em arXiv preprint arXiv:1412.6980}, 2014.

\bibitem{otsu1975threshold}
Nobuyuki Otsu et~al.,
\newblock ``A threshold selection method from gray-level histograms,''
\newblock {\em Automatica}, vol. 11, no. 285-296, pp. 23--27, 1975.

\bibitem{simpson2019large}
Amber~L Simpson, Michela Antonelli, Spyridon Bakas, Michel Bilello, Keyvan Farahani, Bram Van~Ginneken, Annette Kopp-Schneider, Bennett~A Landman, Geert Litjens, Bjoern Menze, et~al.,
\newblock ``A large annotated medical image dataset for the development and evaluation of segmentation algorithms,''
\newblock {\em arXiv preprint arXiv:1902.09063}, 2019.

\bibitem{beichel2004liver}
Reinhard Beichel, Thomas Pock, Christian Janko, Roman~B Zotter, Bernhard Reitinger, Alexander Bornik, Kalman Palagyi, Erich Sorantin, Georg Werkgartner, Horst Bischof, et~al.,
\newblock ``Liver segment approximation in ct data for surgical resection planning,''
\newblock in {\em Medical Imaging 2004: Image Processing}. SPIE, 2004, vol. 5370, pp. 1435--1446.

\bibitem{sun2020vascular}
Mengyu Sun, Wenjuan Lv, Xinyan Zhao, Lili Qin, Yuqing Zhao, Xiaohong Xin, Jianbo Jian, Xiaodong Chen, and Chunhong Hu,
\newblock ``Vascular branching geometry relating to portal hypertension: a study of liver microvasculature in cirrhotic rats by x-ray phase-contrast computed tomography,''
\newblock {\em Quantitative imaging in medicine and surgery}, vol. 10, no. 1, pp. 116, 2020.

\bibitem{wolny2020accurate}
Adrian Wolny, Lorenzo Cerrone, Athul Vijayan, Rachele Tofanelli, Amaya~Vilches Barro, Marion Louveaux, Christian Wenzl, S{\"o}ren Strauss, David Wilson-S{\'a}nchez, Rena Lymbouridou, et~al.,
\newblock ``Accurate and versatile 3d segmentation of plant tissues at cellular resolution,''
\newblock {\em Elife}, vol. 9, pp. e57613, 2020.

\end{thebibliography}
\end{document}